\documentclass{article} 
\usepackage{iclr2016_workshop,times}
\usepackage{hyperref}
\usepackage{url}
\usepackage{times}
\usepackage{helvet}
\usepackage{courier}
\usepackage{url}
\usepackage{graphicx}
\usepackage{caption}
\usepackage{subcaption}
\usepackage{placeins}
\usepackage{multirow}
\usepackage{wrapfig}
\usepackage{footnote}
\usepackage{tablefootnote}
\usepackage{amssymb}
\usepackage{amsmath}
\usepackage{amsfonts}
\usepackage{lipsum}         
\usepackage{float}          

\title{Robust Convolutional Neural Networks \\ under Adversarial Noise}

\author{Jonghoon Jin, Aysegul Dundar and Eugenio Culurciello\\
  Electrical and Computer Engineering, Purdue University\\
  Biomedical Engineering, Purdue University\\
  \texttt{\{jhjin,adundar,euge\}@purdue.edu} \\
}

\begin{document}

\maketitle

\begin{abstract}
  Recent studies have shown that Convolutional Neural Networks (CNNs) are vulnerable to a small perturbation of input called ``adversarial examples''.
  In this work, we propose a new feedforward CNN that improves robustness in the presence of adversarial noise.
  Our model uses stochastic additive noise added to the input image and to the CNN models.
  The proposed model operates in conjunction with a CNN trained with either standard or adversarial objective function.
  In particular, convolution, max-pooling, and ReLU layers are modified to benefit from the noise model.
  Our feedforward model is parameterized by only a mean and variance per pixel which simplifies computations and makes our method scalable to a deep architecture.
  From CIFAR-10 and ImageNet test, the proposed model outperforms other methods and the improvement is more evident for difficult classification tasks or stronger adversarial noise.
\end{abstract}

\section{Introduction}

Convolutional neural networks (CNNs) \citep{lecun1998gradient} have shown great success in visual and semantic understanding.
They have been applied to solve visual recognition problems, where hard-to-describe objects or multiple semantic concepts are present in images.

Given the global widespread use of cameras on mobile phones, CNNs are already a candidate to perform categorization of user photos.
Device manufacturers use various types of cameras, each with very different sensor noise statistics \citep{tian2000noise}.
Also, recent phone cameras can record a video at hundreds of frames per second, where more frames per second translates into higher image noise \citep{tian2000noise}.
Unfortunately, CNNs are vulnerable to artificial noise and could be easily fooled by the noise of just few pixels \citep{szegedy2013intriguing,goodfellow2014explaining,nguyen2014deep}.
This problem arises because standard CNNs are discriminative models. 
This work provides a solution to improve instability of CNNs, for example in security applications.

While \citet{bishop1995training} showed that training with noise is equivalent to a regularizer,
\citet{goodfellow2014explaining,huang2015learning} used adversarial perturbation during training but not has been applied to natural images or more challenging image classification tasks.
Similarly, \citet{gu2014towards} proposed a denoising model for adversary using auto-encoders.

The main contribution of this work is to propose a robust feedforward CNN model under adversarial noise; a noise that affects the performance the most.
In order to achieve this, we add stochasticity to the CNN models with the assumption that the perturbation can be seen as a sample drawn from a white Gaussian noise.
Our model takes advantage of a parametric model, which makes our method possible to scale up and apply to a large-scale dataset such as ImageNet.

\section{Convolutional Neural Networks with Noise Model}

The proposed feedforward model uses a noise distribution applied to each pixel.
The following subsections explain the stochastic operation of each layer, including convolution, pooling, and non-linear activation.
Also, the rest of operators used in standard CNNs can be found in appendix \ref{appendix_operator}.

\subsection{Input Noise Model}

We add an uncertainty to input images so the CNN output in hyperspace becomes a cloud with uncertainty information instead of a vector.
We hypothesize that referring to the marginal data during classification helps CNNs to be robust toward adversarial examples.
As a result of our modeling, each pixel becomes a random variable $X$ in $\mathbb{R}^3$ (channel, height, width) and follows a normal distribution with the mean of the original pixel value $\mu_{X_{ijk}}$ and a constant variance of $\sigma_N^2$
\begin{equation}
  \begin{aligned}
    X_{ijk} \triangleq \mu_{X_{ijk}} + N \Longrightarrow X_{ijk} \sim \mathbf{N}(\mu_{X_{ijk}}, \sigma_N^2) \\
  \end{aligned}
\end{equation}
where all input pixels have the same noise power of $\sigma_N^2$.
The conditional independence of noise, for given value $\mu_{X_{ijk}}$, among pixels in neighborhood helps us to simplify the model and make it scalable to deep networks.
To clarify, we adopted the artificial noise distribution in order to improve the robustness of CNNs and it is unrelated to natural image statistics.

\subsection{Convolution Layer}

While CNN inputs are modeled as random variables, all remaining parameters such as weights and biases are fixed constants. 
Convolution is a weighted sum of random variables and its first and second order moments of output of convolution layer are shown in equation \ref{eq:conv0}.
\begin{equation}
  \begin{aligned}
    E\left[Y\right]   = & \: \sum \omega E\left[X\right] + b, \qquad
    Var\left[Y\right] = & \: \sum \omega^2 Var\left[X\right] \\
  \end{aligned}
  \label{eq:conv0}
\end{equation}
$X$ and $Y$ corresponds to a single element of the input and output in the convolution layer respectively.
$\omega$ and $b$ are weights and biases, and pixel index $i$, $j$ and $k$ are omitted for conciseness.
We are interested in the first and second order statistics since we want to stay with a parametric model throughout layers, which simplifies overall computations.

\subsection{Rectified Linear Unit Layer}

Rectified linear units (ReLU) \citep{krizhevsky2012imagenet} applies the non-linearity $Y = max(X, \theta)$ in an element-wise manner.
Then, as illustrated in appendix \ref{appendix_dist}, the distribution of the stochastic ReLU output $Y$ is left-censored where $Y = X$ for $X > \theta$ otherwise reported as a single value $\theta$.
The mean and variance \citep{greene2008econometric} of output $Y$ for the given normal distribution of input $X$ are:
\begin{equation}
  \begin{aligned}
    E\left[Y\right]   = & \: E[Y|X=\theta] Pr(Y=\theta|X) + E[Y|X>\theta] Pr(Y>\theta|X) \\
                      = & \: \theta \Phi(\alpha) + \left(\mu_X + \sigma_X \lambda(\alpha)\right) \left(1 - \Phi(\alpha)\right) \\
    Var\left[Y\right] = & \: E_X[Var[Y|X]] + Var_X[E[Y|X]] \\
                      = & \: \sigma_X^2 (1\!-\!\Phi(\alpha)) \! \left[ (1\!-\!\delta(\alpha)) \!+\! (\alpha \!-\! \lambda(\alpha))^2 \Phi(\alpha) \right]
  \end{aligned}
  \label{eq:threshold0}
\end{equation}
where $\delta(\alpha) = \lambda(\alpha)(\lambda(\alpha) - \alpha)$, $\lambda(\alpha) = \phi(\alpha) / (1-\Phi(\alpha))$, a standard score $\alpha = (\theta - \mu_X)/\sigma_X$, a standard normal density of $\phi$ and a cumulative normal distribution of $\Phi$ are used.
Outputs from the convolution layer followed by non-linearity are reasonably approximated to be Gaussians by the central limit theorem
considering that an output neuron has more than few hundreds of connections in general.
Stochastic ReLU operator allows to deliver tail information of the distribution to the higher layer of CNNs regardless of the neuron's activation, which is expected to contribute better decision making.

\subsection{Max-pooling Layer}
\label{sec:maxpool}

Prediction from stochastic max-pooling is calculated based on the exact distribution of the max of two normal distributions \citep{nadarajah2008exact}
whose variables are sampled from a set $S$ with elements in the pooling window.
The pairwise max operation in the equation \ref{eq:maxpool0} is iteratively applied until no element left in the set $S$.
\begin{equation}
  \begin{aligned}
    E[Y] = & \: \mu_{X_i} \Phi \left( \alpha \right) +
                \mu_{X_j} \Phi \left(-\alpha \right) +
                \theta \phi \left( \alpha \right) \\
    Var[Y] = & \: (\sigma_{X_i}^2 + \mu_{X_i}^2) \Phi \left( \alpha \right) +
                  (\sigma_{X_j}^2 + \mu_{X_j}^2) \Phi \left(-\alpha \right) + (\mu_{X_i} + \mu_{X_j}) \theta \phi \left( \alpha \right) - E[Y]^2
  \end{aligned}
  \label{eq:maxpool0}
\end{equation}
where $\alpha = \frac{(\mu_{X_i} - \mu_{X_j})}{\theta}$, $\theta = \sqrt{\sigma_{X_i}^2 + \sigma_{X_j}^2 }$.
By approximating the output to a normal parametric distribution (see appendix \ref{appendix_dist}), we trade off accuracy for the sake of scalability of this method.
According to \citet{sinha2007advances} and appendix \ref{appendix_approx}, the ordering of iterative max operation should be set in an ascending order by their means to minimize approximation error.

\section{Experimental Results}

\begin{table}[t!]
  \vspace{-23pt}
  \caption{
    Classification accuracy on CIFAR-10 and ImageNet validation sets.
    The adversarial training with high noise on ImageNet failed to converge, which is marked with an asterisk.
    The zero adversarial noise intensity corresponds to the original validation set.
    A different input variance $\sigma_N^2$ in stochastic model was used for each level of adversarial noise.
  }
  \vspace{-7pt}
  \label{table-result}
  \begin{center}
    \begin{tabular}{l|rr|rrr}
      \multicolumn{1}{c}{\bf }  &\multicolumn{2}{|c|}{\bf CIFAR-10 (\%)}  &\multicolumn{3}{|c}{\bf ImageNet (\%)} \\
      \multicolumn{1}{c}{\bf Adversarial noise intensity (k) [px]}
      &\multicolumn{1}{|r}{\bf \qquad 0}  &\multicolumn{1}{r|}{\bf \quad 0.5}
      &\multicolumn{1}{|r}{\bf \qquad 0}  &\multicolumn{1}{r}{\bf \quad 0.01}  &\multicolumn{1}{r}{\bf \quad 0.5} \\ \hline \hline
      Standard training (baseline)                          &     90.1 &     72.3 &     57.0 &     56.1 &     24.8 \\ 
      Standard training + Stochastic FF                     &     88.9 & \bf{78.1}&     57.0 & \bf{56.2}& \bf{33.4}\\ 
      \hline                                                                                             
      LWA + BN \citep{huang2015learning}                    &     89.0 &     82.3 &      --- &      --- &      --- \\ 
      Adversarial training \citep{goodfellow2014explaining} &     88.7 &     82.1 &     43.0 &     42.9 &        * \\ 
      Adversarial training + Stochastic FF                  &     88.7 & \bf{82.9}&     43.0 &     42.9 &        * \\ 
    \end{tabular}
  \end{center}
  \vspace{-10pt}
\end{table}

We denote the proposed method as ``stochastic feedforward (FF)'' throughout the section.
The stochastic FF was applied to the Network-in-Network (NIN) \citep{lin2013network} and the single column AlexNet \citep{krizhevsky2014one} with the latest technique including dropout and batch normalization \citep{ioffe2015batch}.
The networks were trained with either standard or adversarial training \citep{goodfellow2014explaining} for comparison.
Their performance under adversarial noise was evaluated on CIFAR-10 and ImageNet classification datasets \citep{krizhevsky2009learning,russakovsky2014imagenet}.
The gradient sign method \citep{goodfellow2014explaining} was used to generate adversarial examples that are more likely to appear in natural environment.

The table \ref{table-result} summarizes the best classification results obtained from our experiments.
In the presense of adversarial noise, the stochastic FF provides extra accuracy gain regardless of training methods at the cost of little accuracy drop in normal configuration --- no added noise.
The gain is more visible when the task is difficult or the noise is stronger.
In the ImageNet test, adversarial training converges to a lower accuracy than the standard training,
but also the adversarial training with a noise intensity higher than $0.1$ fails to reach to a meaningful score; better than random guessing.
Considering a byte-precision pixel range $[0, 255]$, the adversarial training is limited for use in a hard task whereas the stochastic FF combined with standard training is still effective in high noise condition.

CNNs' decision boundary is constructed based on sparsely populated training samples \citep{nguyen2014deep} in a high-dimension.
Adversarial examples used in this experiment are populated around the decision boundary,
therefore, they are often indistinguishable from natural images if one uses point-wise prediction.
In the stochastic FF, uncertainty around the input pixel is propagated throughout every layer of CNNs and provides marginal information.
Instead of point-wise prediction, integrating such information increases a chance to make correct prediction for adversarial examples.
Adding stronger noise drags the adversarial examples farther apart from the correct decision region
thereby lowering the accuracy as in appendix \ref{appendix_result}.

A downside of the stochastic FF is accuracy loss due to mistuned input variance or numerical instability.
For example, an input distribution with high variance makes it more likely to be uniform where the ambiguity causes performance degradation.
Users will have to make a choice whether to prefer small loss of classification for large gain in the presence of noise.
For near-zero variance, each pixel distribution is shaped to a Dirac delta function.
Without uncertainty, the model is equivalent to the standard feedforward while near-zero division causes numerical instability.

The stochastic model may be ensembled with the standard CNN model in order to compensate its weakness under the absence of adversarial noise.
the ensemble model on ImageNet is more robust to adversarial noise than the baseline model by $13.12\%$ but with only $0.28\%$ of accuracy loss under normal configuration (appendix \ref{appendix_result}).

\section{Conclusion}

We present new feedforward CNN with stochastic input model that is robust to the adversarial noise.
The proposed model outperforms other methods in the literature and the accuracy gain becomes more evident for difficult classification task or stronger adversarial noise.
Our model takes advantage of a parametric model, which makes our method scalable to a deep architecture like AlexNet.
This work provides a solution how to overcome CNNs' sensitivity to the adversarial noise so as to avoid potential security problems in CNN applications.

\bibliography{stochastic-cnn}
\bibliographystyle{iclr2016_workshop}

\newpage
\appendix

\section*{Appendix to \\ Robust Convolutional Neural Networks \\ under Adversarial Noise}
\renewcommand{\thesubsection}{\Alph{subsection}}

\subsection{Input-Output Distribution}
\label{appendix_dist}

\begin{figure}[H]
  \centering
  \begin{subfigure}[b]{0.5\textwidth}
    \includegraphics[width=\textwidth]{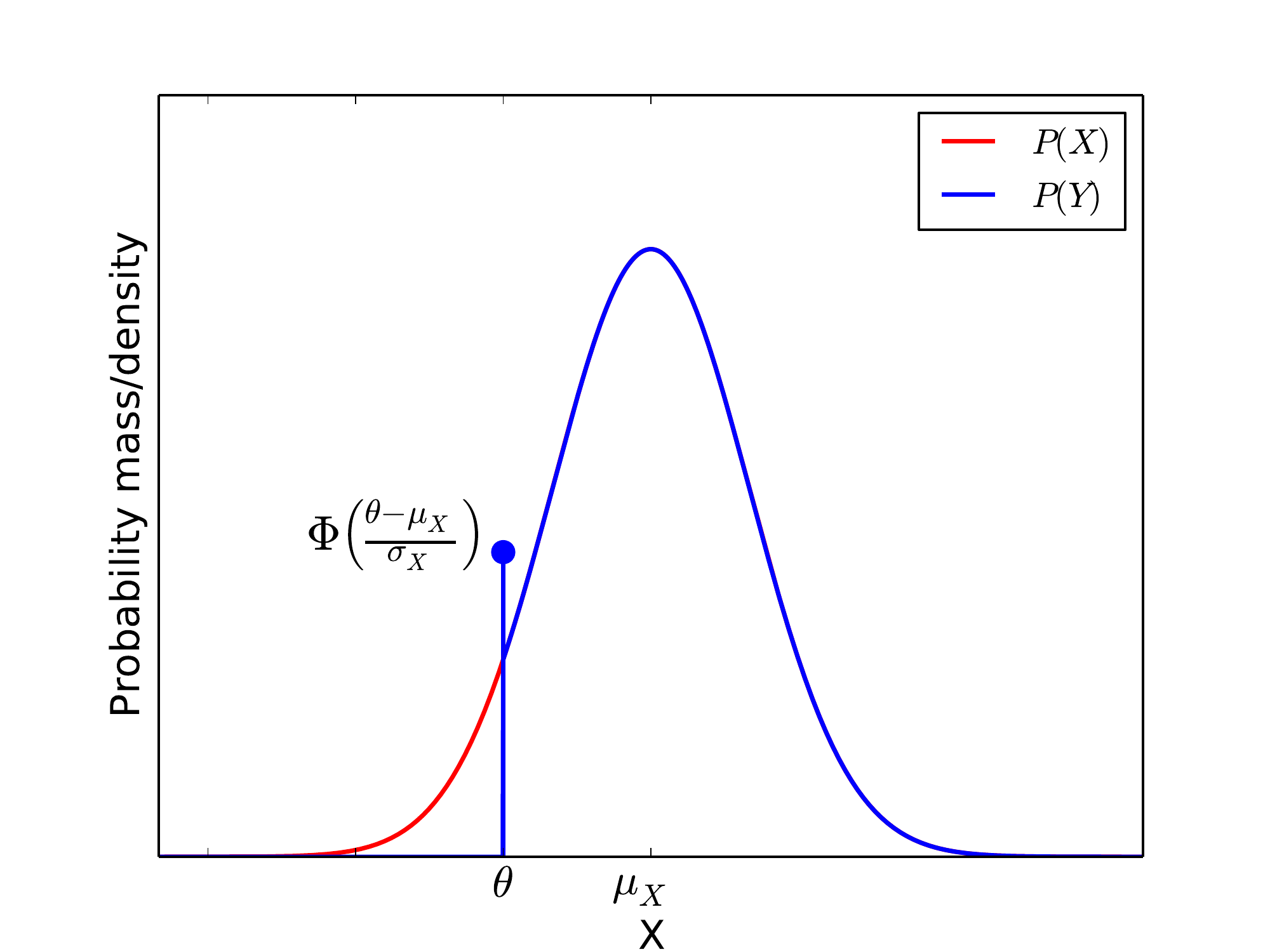}
    \caption{Rectified linear unit (ReLU) layer}
    \label{fig:threshold}
  \end{subfigure}%
  \begin{subfigure}[b]{0.5\textwidth}
    \includegraphics[width=\textwidth]{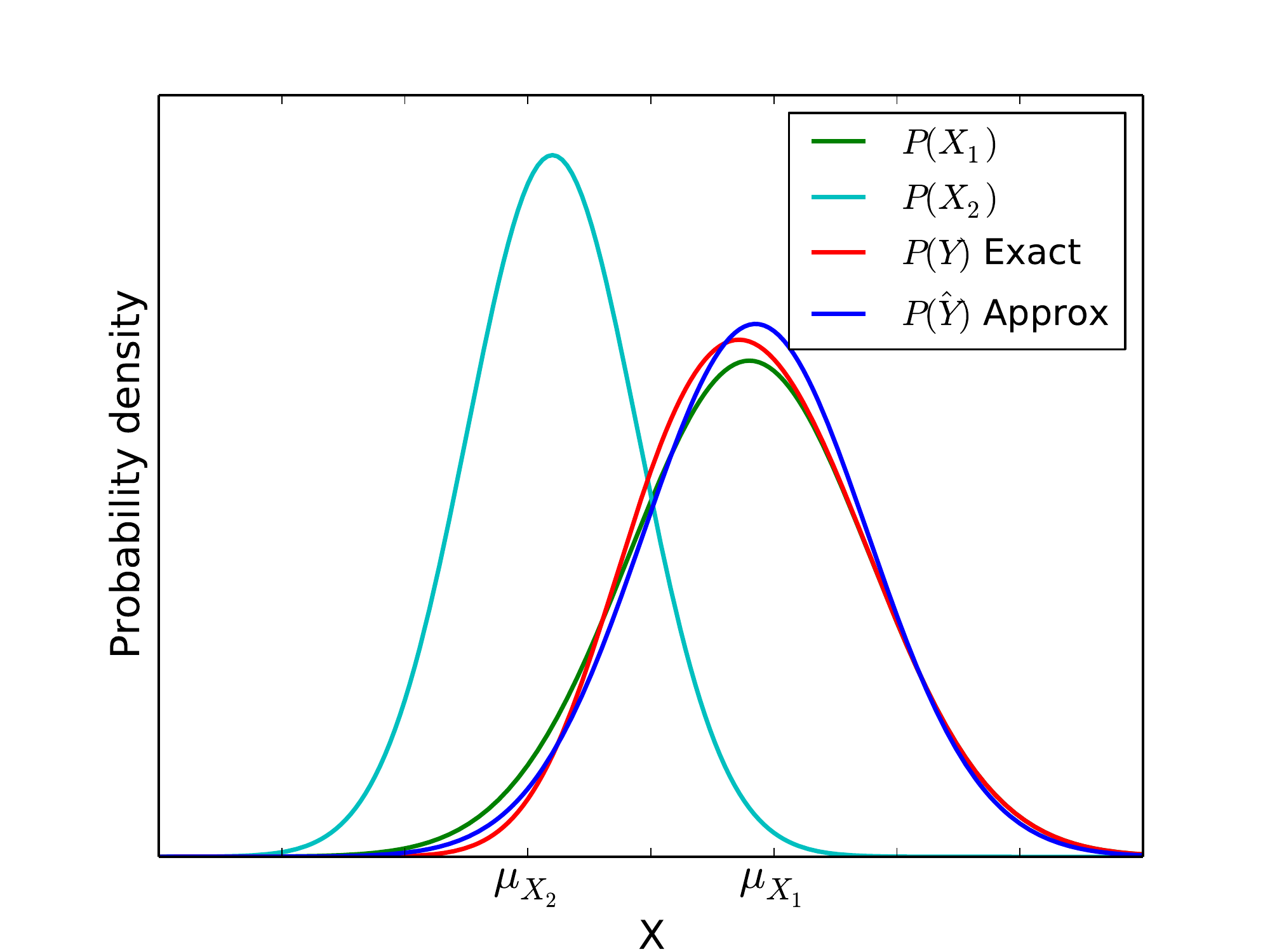}
    \caption{Max-pooling layer}
    \label{fig:maxpool}
  \end{subfigure}
  \caption{
    Behavior of ReLU and max-pooling layer with an example of two stochastic input random variables.
    All input distributions are assumed to be normally distributed.
    (a) The area $\Phi\left(\cdot\right)$ under the red curve between $\left(-\infty, \theta\right]$ is left-censored where its area is reported as a probability mass of $Y$ at the threshold $\theta$.
    (b) The curves in red and blue illustrate the exact distribution $P(Y)$ of the max of two input random variables denoted as $X_1$, $X_2$ centered at $\mu_{X_1}$, $\mu_{X_2}$, and its normal approximation $P(\hat{Y})$ calculated from equation \ref{eq:maxpool0}.
  }
  \label{fig:distribution}
\end{figure}

An example in the figure \ref{fig:threshold} illustrates the input and output distributions from a single neuron in the ReLU layer,
where the stem at $\theta$ indicates point mass probability for the deactivated area with respect to the threshold $\theta$.
The output from the ReLU layer is a censored distribution whose approximation causes error during feedforward computation.

The figure \ref{fig:maxpool} illustrates the input, output and approximated output distributions from max-pooling layer simulated with only two variables.
The output distribution is bell-shaped, but its mode is inclined to the right.
When means of the two inputs are farther, the resulting distribution is likely to be a normal distribution.
In other words, the error between the exact distribution $P(Y)$ and its approximated Gaussian $P(\hat{Y})$ tends to increase as the difference of means increases.

\subsection{Other stochastic operations}
\label{appendix_operator}

Along with the main operations previously discussed, standard CNNs consist of other modules such as batch normalization \citep{ioffe2015batch}, spatial average pooling \citep{lin2013network}, softmax (or log-softmax) and dropout \citep{hinton2012improving}.

The spatial average pooling and the batch normalization are linear functions and they can be processed with the convolution layer model.
The average pooling used in NIN architecture is equivalent to convolution layer whose weights and biases are replaced with averaging coefficients $(1/n)$  and zeroes respectively.
Also, the net operation of batch normalization in evaluation phase is an affine transformation (equation \ref{eq:batchnorm}) where $\gamma$ and $\beta$ are constants learned from training,
therefore, the same convolution modeling can be applied to here without extra approximation.
\begin{equation}
  \begin{aligned}
	  Y = \gamma \frac{(X - \mu_{X})}{\sqrt{\sigma^2_{X} + \epsilon}} + \beta
  \end{aligned}
  \label{eq:batchnorm}
\end{equation}

The softmax with deterministic input produces pseudo-probabilities whose highest activation predicts class category.
The proposed method adopts Gaussians to model intermediate representations throughout the network.
The strongest activation among all class distributions can be easily distinguished by the mean values of Gaussians.
Therefore, we process mean values without variances as like in the normal softmax layer.
Dropout neurons are simply deactivated and work as identity functions during evaluation.

\subsection{Parameter tuning}

\subsubsection{Adversarial Examples}

Prior to the experiment, adversarial examples are being generated through the fast gradient sign method proposed in \citet{goodfellow2014explaining}.
The method generates adversarial noise on top of natural images whose direction is toward the opposite of a gradient.
Though the difference between the original and adversarial sample is imperceptible to human eyes,
the perturbation makes the samples cross the decision boundary therefore classified as different categories.
The noise can be interpreted as an exceptional type of noise although such examples are hardly observed in the natural environment.

Considering that the pixels encoded in an 8-bit image are positive integers in the range of $[0, 255]$,
the smallest and effective pixel intensity of the sign should be a multiple of $1/\sigma_C$ where $\sigma_C$ is a standard deviation used for channel-wise normalization during data preprocessing.
We denote $k_{adv}/\sigma_C$ as a normalized intensity of adversarial noise and only tune the pixel intensity $k_{adv}$ throughout all experiments.
However, we used a continuous range along with effective intensities in byte representation so that input-output relation is more apparent and easily observable.

\subsubsection{Input Variance Tuning}

Upon the creation of stochastic input model, we need to choose a variance for input distributions.
The uncertainty variable (or input variance) is the only parameter in this model.
The stochastic input is artificial modeling and it is designed to take into account the possible range of adversarial noise.
Therefore, it needs to be tuned differently based on the intensity of adversarial noise.

Using very small variance for input makes the method mathematically equivalent to the baseline CNN model.
However, both ReLU and max-pooling layer with the stochastic input model requires division, which is exposed to numerical instability from near-zero denominators.
We found $\epsilon$ of $1e^{-20}$ minimized the numerical error.
The number is adopted for regularizing the denominators both on CIFAR-10 and ImageNet.

\begin{figure}[H]
  \centering
  \begin{subfigure}[b]{0.5\textwidth}
    \includegraphics[width=\textwidth]{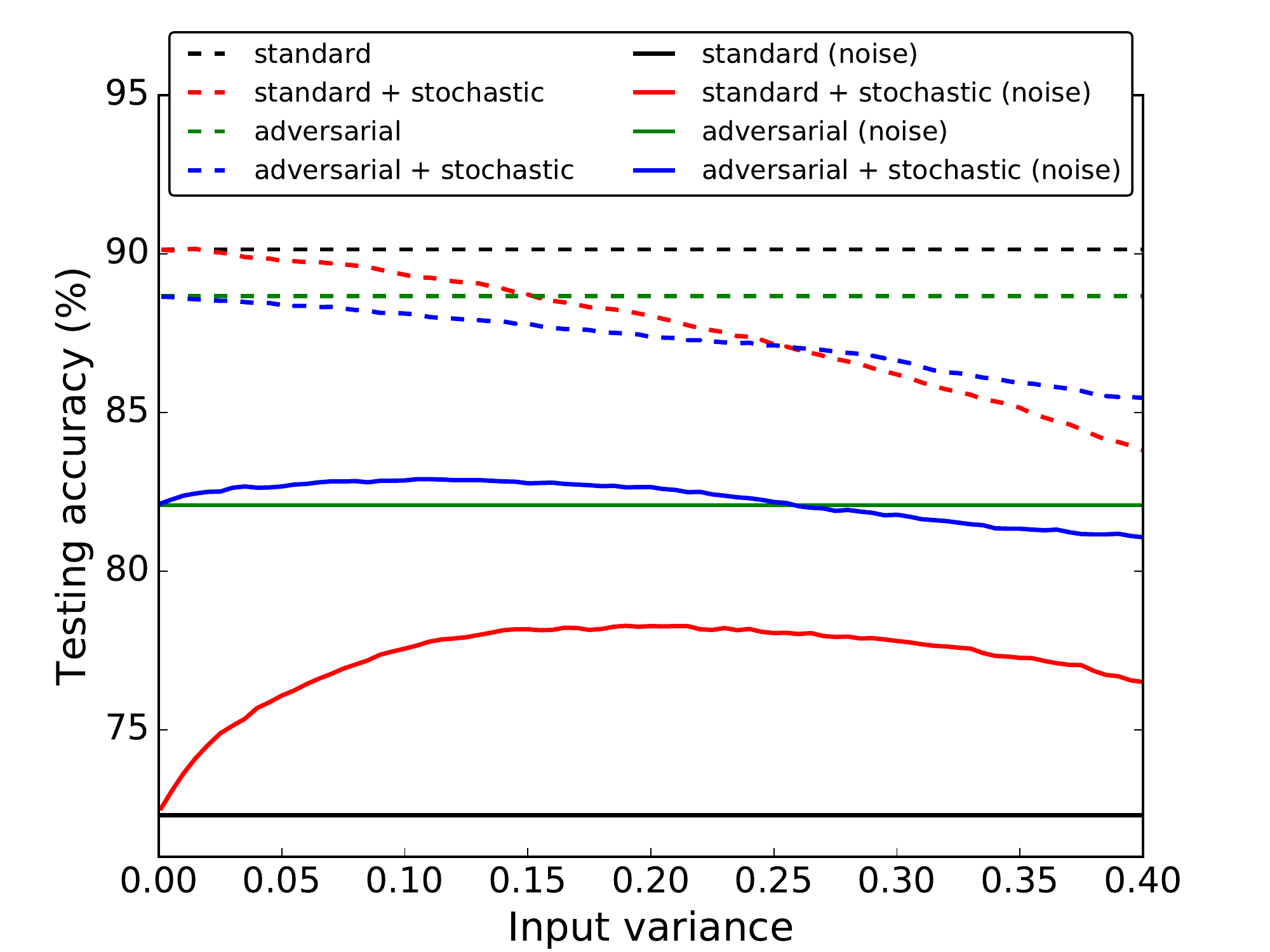}
    \caption{CIFAR-10 test with $k_{adv} = 0.5$}
    \label{fig:cifar10-acc-vs-var}
  \end{subfigure}%
  \begin{subfigure}[b]{0.5\textwidth}
    \includegraphics[width=\textwidth]{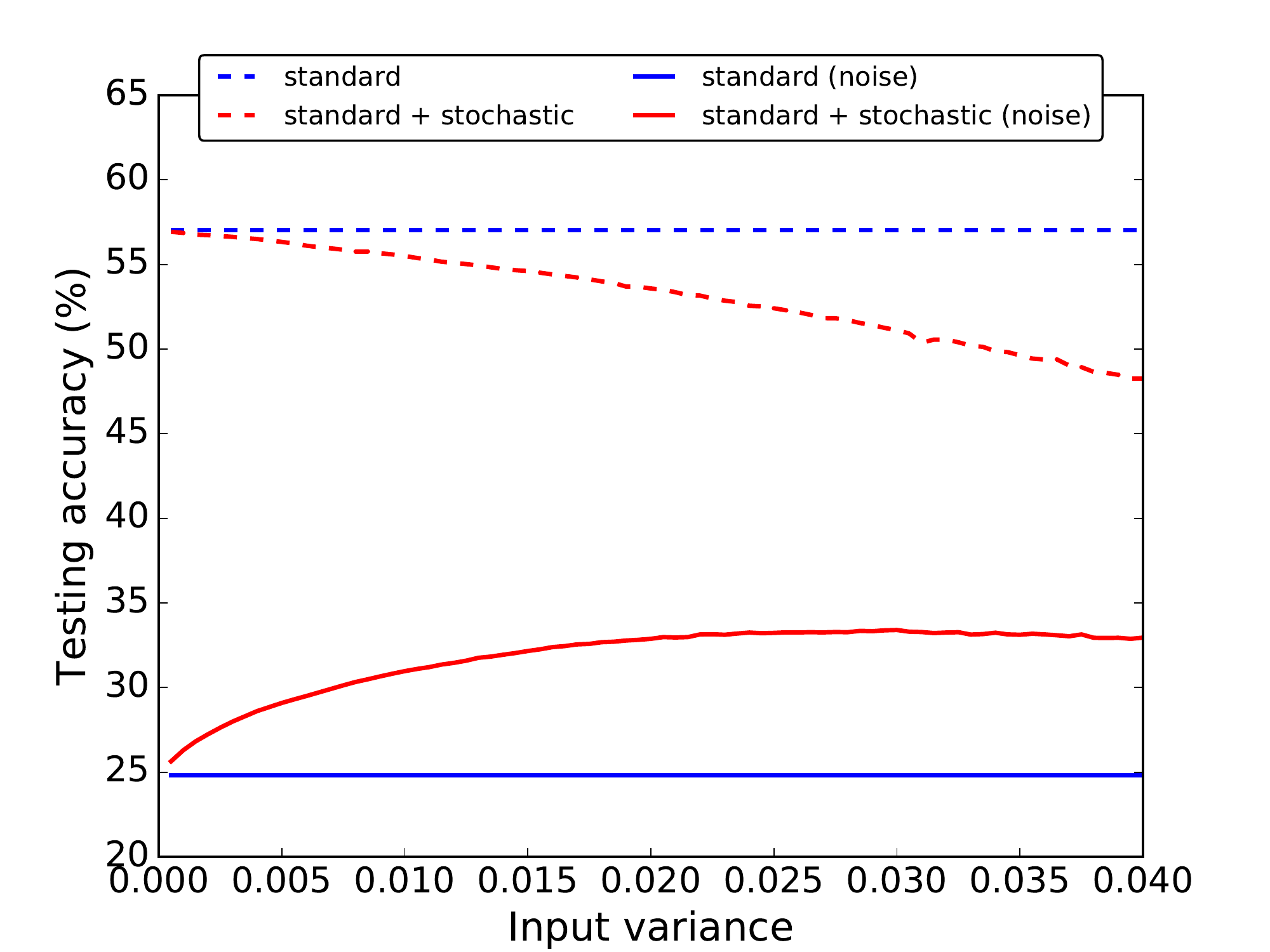}
    \caption{ImageNet test with $k_{adv} = 0.5$}
    \label{fig:imagenet-acc-vs-var}
  \end{subfigure}
  \caption{
    Performance versus different input variance ($\sigma_N^2$) on CIFAR-10 and ImageNet dataset for a fixed adversarial noise intensity.
    The NIN and the single column AlexNet were used in the experiment.
  }
  \label{fig:performance-var}
\end{figure}

\subsection{More Results}
\label{appendix_result}

\begin{figure}[H]
  \centering
  \begin{subfigure}[b]{0.5\textwidth}
    \includegraphics[width=\textwidth]{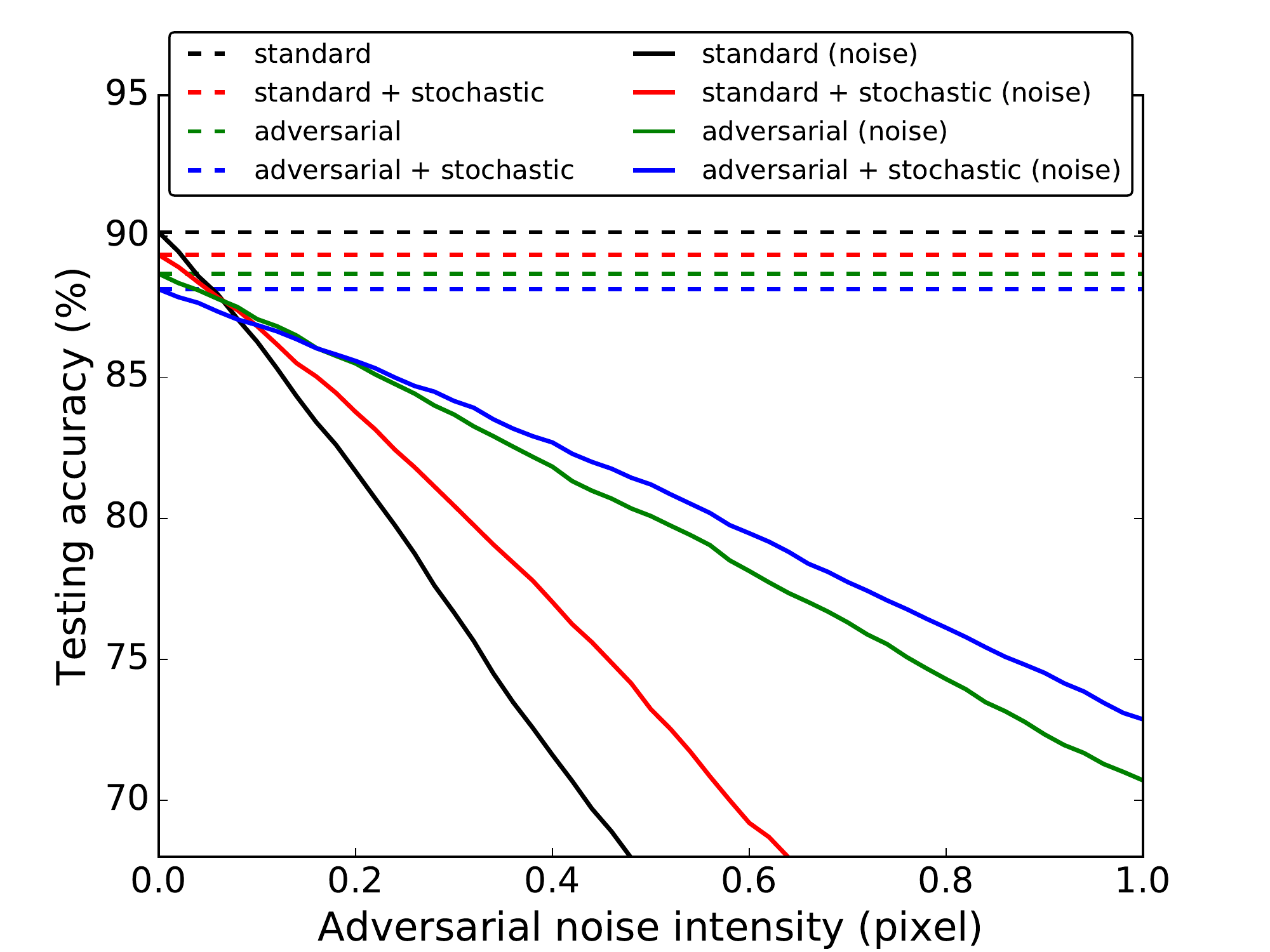}
    \caption{CIFAR-10 test with $\sigma_N^2 = 0.1$}
    \label{fig:cifar10-acc-vs-adv}
  \end{subfigure}%
  \begin{subfigure}[b]{0.5\textwidth}
    \includegraphics[width=\textwidth]{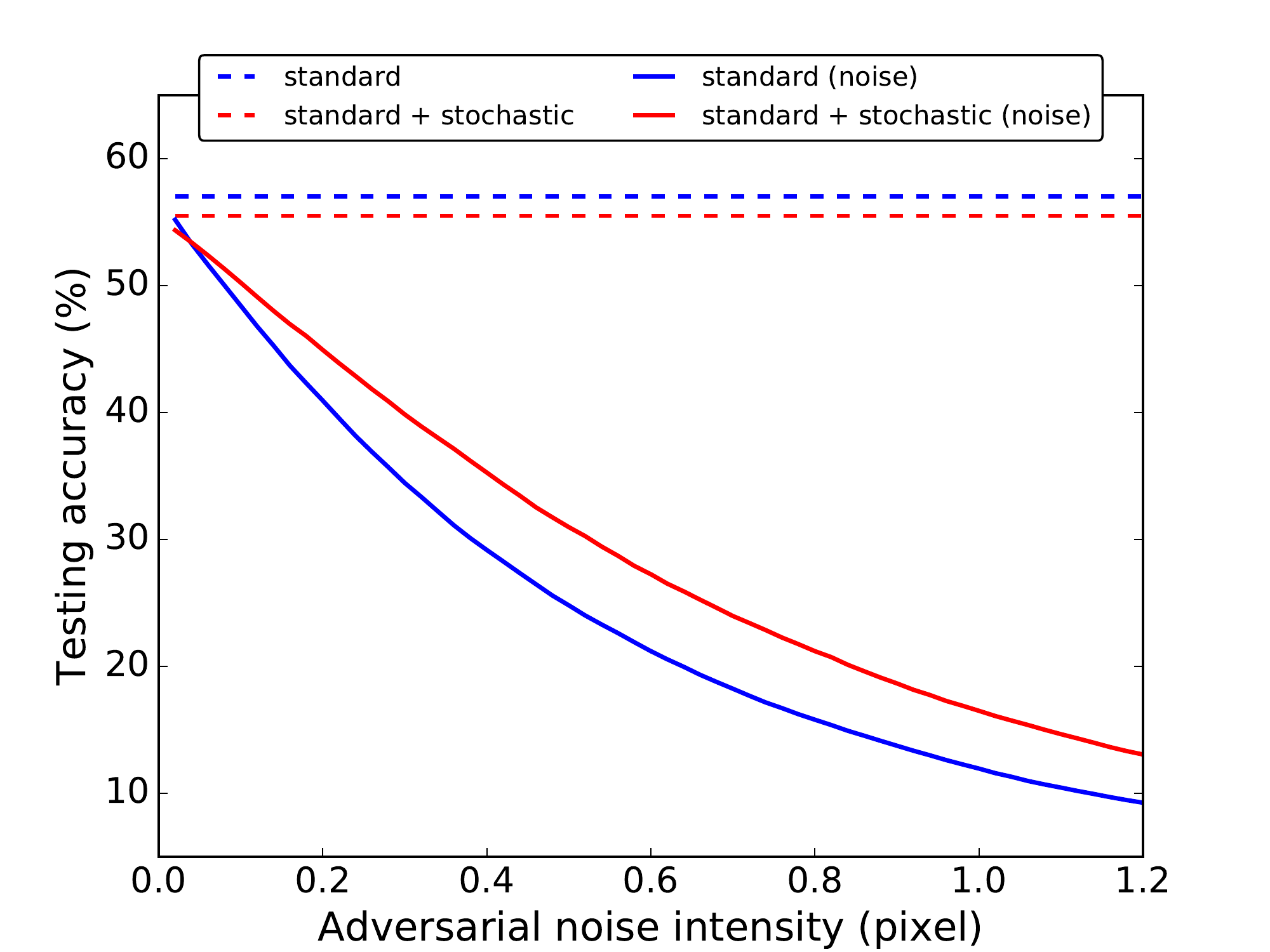}
    \caption{ImageNet test with $\sigma_N^2 = 0.01$}
    \label{fig:imagenet-acc-vs-adv}
  \end{subfigure}
  \caption{
    Performance versus different adversarial noise intensity ($k_{adv}$) on CIFAR-10 and ImageNet dataset for a fixed input variance.
    The NIN and the single column AlexNet were used in the experiment.
  }
  \label{fig:performance-adv}
\end{figure}

\begin{figure}[H]
  \centering
  \includegraphics[width=0.5\textwidth]{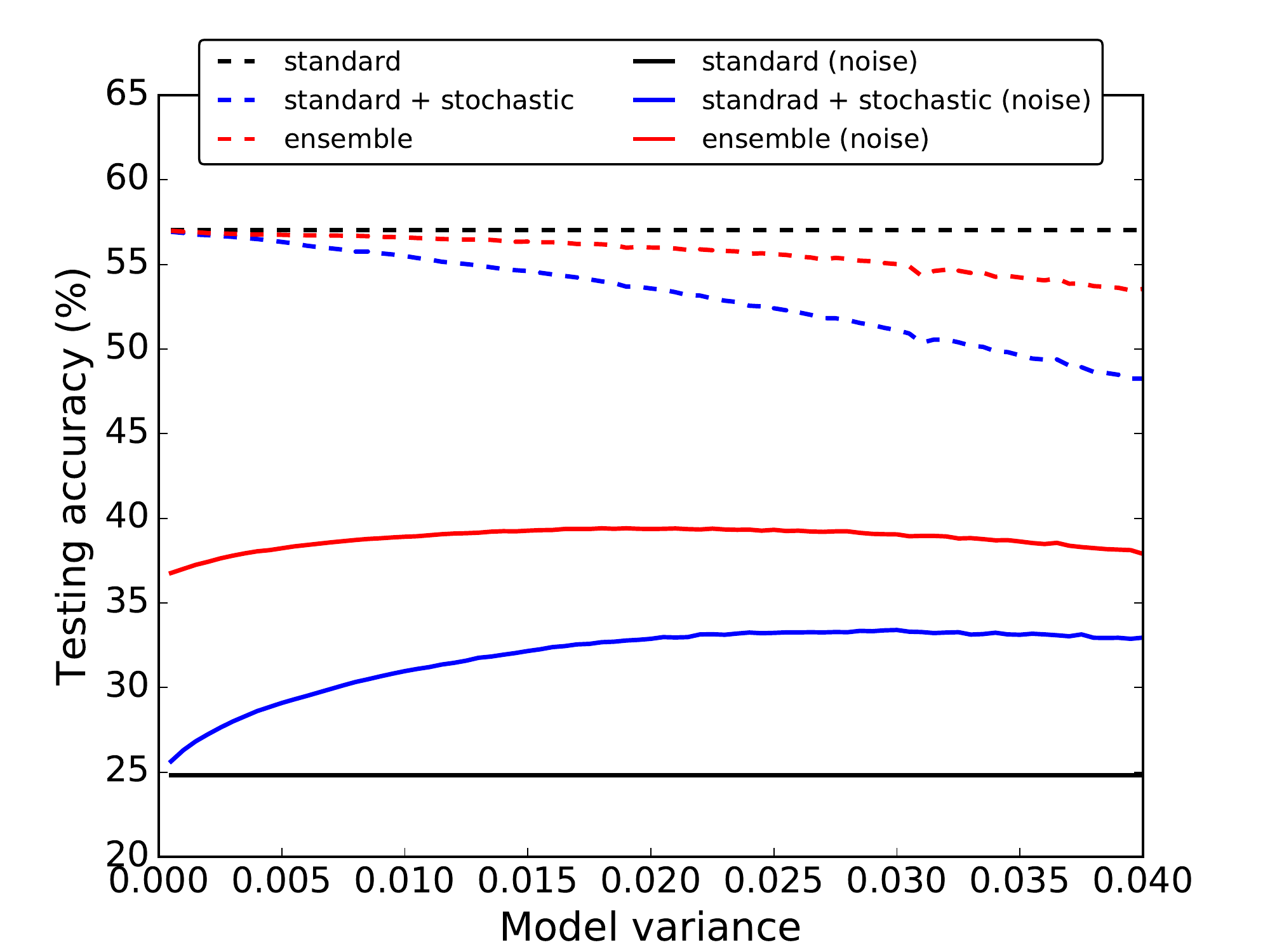}
  \caption{
    Performance of ensemble models over different input variance ($\sigma_N^2$) with $k_{adv} = 0.5$ in the presence of adversarial noise on ImageNet.
    The output probabilities from the standard model and standard model with stochastic FF are averaged with a ratio of 1:1.
  }
  \label{fig:ensemble-imagenet}
\end{figure}

\subsection{Approximation Error}
\label{appendix_approx}

The stochastic method based on a parametric model simplified computation,
but it accompanies approximation in max-pooling and ReLU layer, which is a limiting factor of the algorithm.

\begin{figure}[H]
  \centering
  \includegraphics[width=0.5\textwidth]{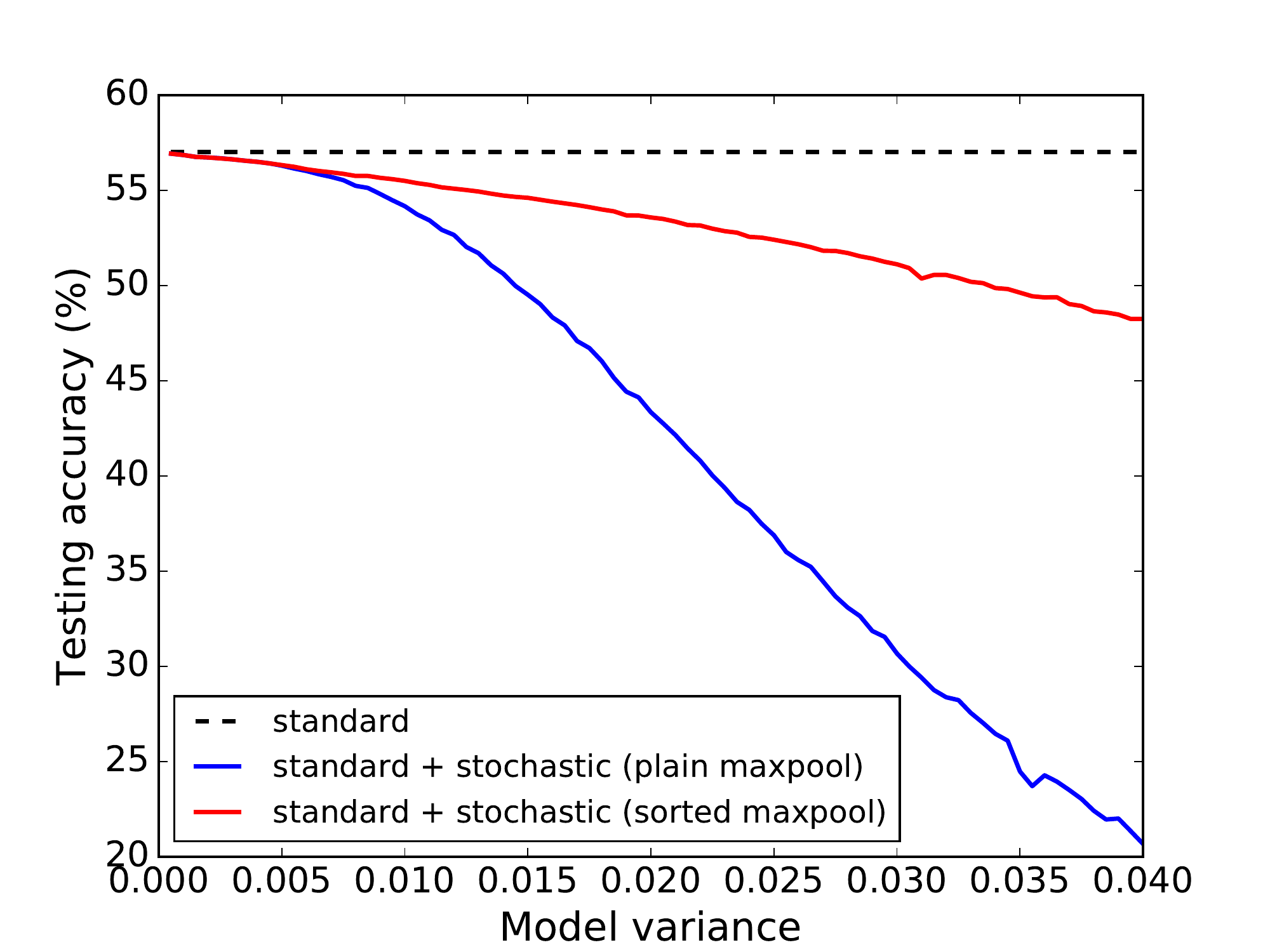}
  \caption{
    The accuracy gap due to the approximation error between the baseline and stochastic AlexNet without any perturbation.
    Sorted max-pooling achieves lower approximation error than plain max-pooling especially for the large variance ($\sigma_N^2$).
  }
  \label{fig:maxpool-approx}
\end{figure}

From the earlier section, it is expected that the max-pooling with stochastic input model produces approximation error and its quantity depends on the ordering of elements.
We tested accuracy due to the approximation error from the plain and ordered max-pooling to check the effectiveness of sorting in the context of CNNs.
A single column AlexNet has more max-pooling layers with a larger pooling region than the NIN.
Therefore, it is more vulnerable to approximation error and used to observe the effectiveness as reported in figure \ref{fig:maxpool-approx}.
We found that the sorted max-pooling gave higher accuracy most of time compared to the plain order.
The plain max-pooling fails to draw accurate approximation toward the exact distribution for a high variance model due to its heavy-tailed distribution.
In terms of computation, the cost of the sorting is negligible considering the fact that the pooling is employed at most on an $3\times3$ array in the literature \citep{krizhevsky2012imagenet}.

The ReLU operator often suffers from numerical instability.
Intermediate representations in CNNs are generally sparse meaning that many values are populated around zero.
From the equation \ref{eq:threshold0}, these near-zero values combined with tiny input variance produce non-trivial standard scores, which requires regularization.
This regularization process makes the approximated value deviate from its true value.

In general, as CNNs become deeper, they are also sensitive to error accumulated during feedforward computation.
Additionally, performance improvement has been achieved at the cost of additional computation and memory space.
In the stochastic model, each input or intermediate value other than model parameters such as weights and biases is represented as a set of mean and variance.
Therefore, it requires about two times of memory usage than the standard feedforward model.



\end{document}